\documentclass{article}
\usepackage{ijcai26}

\usepackage{enumitem}
\usepackage{times}
\usepackage{soul}
\usepackage{url}
\usepackage[hidelinks]{hyperref}
\usepackage[utf8]{inputenc}
\usepackage[small]{caption}
\usepackage{graphicx}
\usepackage{amsmath}
\usepackage{amsthm}
\usepackage{amssymb}
\usepackage{booktabs}
\usepackage{algorithm}
\usepackage{algorithmic}
\usepackage[switch]{lineno}
\usepackage{tabularx}

\title{Directional Hallucinations: Ideological Drift in News-Grounded LLM Question Answering}

\author{
Chendi Wang
\and
Liam Cunningham
\and
Tom Yishay
\and
Jieying Chen
\affiliations
Vrije Universiteit Amsterdam\\
\emails
chendi.wang@vu.nl,
l.e.cunningham@student.vu.nl,
t.yishay3@student.vu.nl,
j.y.chen@vu.nl
}

\begin{document}

\maketitle

\begin{abstract}
    Large language models (LLMs) are increasingly used to answer questions about political information, including in election-adjacent information settings where factual errors and ideological distortions are high-stakes. We present a reproducible measurement framework that treats hallucinations, unsupported statements in document-grounded QA, as diagnostic signals of ideological drift. Using 21,727 expert-labeled U.S. political news articles from QBias spanning left, center, and right sources, we (i) generate an article-specific question, (ii) elicit document-grounded answers from three open-weight LLMs and one proprietary model, (iii) detect sentence-level hallucinations via reference-based comparison, (iv) classify the ideological valence of hallucinated sentences with a fine-tuned stance classifier, and (v) probe output logits to relate token-level uncertainty to hallucination and drift. Hallucination rates vary substantially across models and concentrate in contentious topics, while source-ideology differences in hallucination frequency are modest. In contrast, hallucination content exhibits robust leftward drift: a majority of hallucinated sentences are classified as left-leaning, including among hallucinations generated from right-leaning sources. Logit-level analysis shows hallucinations arise in high-entropy generation contexts, and in some models uncertainty also predicts leftward drift, consistent with an ``uncertainty $\rightarrow$ guessing'' mechanism. We discuss implications for auditing AI-mediated political information and for designing safeguards in election-relevant deployments. 

\end{abstract}

\section{Societal Problem and Real-World Impact}

\noindent\textbf{Societal problem.}
As LLMs are integrated into search, chat assistants, and news interfaces, they increasingly mediate political information access. In civic and electoral contexts, hallucinated or ideologically skewed answers can misinform voters, distort issue salience, and erode trust in institutions and platforms ~\cite{weidinger2021ethical,bender2021dangers,allcott2017social,vosoughi2018spread}. Recent evidence shows conversational AI can be politically persuasive in controlled settings ~\cite{hackenburg2025levers,lin2025persuading,hackenburg2024evaluating}, raising concerns about downstream effects when AI systems generate incorrect or ideologically slanted claims.

\noindent\textbf{Real-world impact.}
Our output is an auditable, model-agnostic evaluation workflow that can be used by platforms and public institutions to stress-test LLM deployments for civic information. The pipeline produces: (i) hallucination prevalence by source ideology and topic, (ii) directional drift metrics with uncertainty intervals, and (iii) uncertainty-linked risk indicators usable for abstention or warning policies (e.g., ``do not answer when uncertainty exceeds threshold'') ~\cite{kalai2026evaluating}.

\section{Introduction}

LLMs are increasingly deployed as interfaces to political and civic information, often in \emph{grounded} settings where a system is asked to answer a question using a provided document (e.g., a news article) or retrieved sources. Even in such settings, models can produce hallucinations, i.e. unsupported statements that appear confident and coherent ~\cite{huang2025survey,ji2024anah,kalai2026evaluating}. Separately, a growing literature documents political bias and value misalignment in LLM outputs, frequently showing systematic skews that vary by topic and evaluation design ~\cite{rettenbergerAssessingPoliticalBias2025,bang2024measuring,chen2026uncovering,navigli2023biases,motokiAssessingPoliticalBias2025,gallegos2024bias}.

\noindent\textbf{Core question.}What happens when these two failure modes intersect? When an LLM loses grounding and begins to ``fill in'' missing information, is the resulting hallucinated content ideologically \emph{directional}? If so, that would imply hallucinations are not merely reliability failures, but also \emph{bias-revealing events} that surface latent priors.

\noindent\textbf{Why this matters for social good.} In civic contexts (e.g., election information access), an LLM that hallucinates could introduce ideologically slanted misinformation even when the underlying source is ideologically opposed or neutral, undermining fairness, trust, and informed decision-making \cite{weidinger2021ethical,bender2021dangers,allcott2017social}.

\noindent\textbf{Contributions.} Our contributions are fourfold:
\begin{itemize}[leftmargin=1em]
    \item \textbf{Measurement framework:} a transparent, reproducible pipeline for auditing \emph{ideological directionality} of hallucinated content in news-grounded QA.
    \item \textbf{Empirical finding:} hallucination rates vary by model and are modestly higher for left sources, but hallucination content shows robust leftward drift even from right sources.
    \item \textbf{Topical risk profiling:} drift and hallucination vulnerability are amplified on contested topics (including election-relevant issues).
    \item \textbf{Mechanistic evidence:} logit-level uncertainty signals predict hallucination events; uncertainty analyses help distinguish ``guessing'' from source-conditioned generation, and clarify whether drift reflects uncertainty or priors.
\end{itemize}

\section{Related Work}

\paragraph{Political bias in language models.}
Multiple approaches quantify political bias in LLMs, including ideology tests, policy questionnaires, and comparisons to human polling distributions \cite{bang2024measuring,santurkarWhoseOpinionsLanguage2023,motokiAssessingPoliticalBias2025,rozadoMeasuringPoliticalPreferences2025}. Bias is not monolithic: it varies by topic and evaluation protocol \cite{motokiAssessingPoliticalBias2025,navigli2023biases,gallegos2024bias}. These findings motivate bias auditing in \emph{realistic tasks} rather than only prompt-based questionnaires.

\paragraph{Hallucinations and grounded generation.}
Hallucinations in QA and summarization are widely documented \cite{huang2025survey,ji2024anah,zhangSirensSongAI2023}. Recent work argues hallucinations arise partly because training and evaluation reward guessing over abstention \cite{kalai2026evaluating}. Detection methods include human evaluation, entailment-style scoring, and learned annotators \cite{gu2024anah}. Our work uses a reference-based hallucination detector to identify unsupported sentences in grounded QA outputs.

\paragraph{Uncertainty and abstention.}
Confidence estimates in neural models are often miscalibrated \cite{guo2017calibration}. Selective prediction and abstention policies can mitigate harm when uncertainty is high \cite{geifman2017selective}. In LLM contexts, logit-derived metrics (entropy, margins) provide interpretable uncertainty signals that can be connected to hallucination risk \cite{kalai2026evaluating,jiang2021can}.

\paragraph{Democracy, elections, and conversational AI.}
AI systems can influence political attitudes and persuasion dynamics \cite{hackenburg2025levers,lin2025persuading,bai2025llm}. These results heighten the importance of auditing not only overt persuasion, but also \emph{misinformation pathways} produced through hallucinated civic and political information. This question carries significant societal implications. If hallucinations systematically inject political bias into news-grounded QA, users assuming AI neutrality may unknowingly consume ideologically skewed misinformation. Such content could reinforce echo chambers~\cite{EchoChamberEffect2025}, distort public perception, and erode trust in both AI systems and the information they convey, particularly concerning given LLMs' growing role in search and content moderation~\cite{NoelsModeration}. The problem is compounded by the fact that hallucinated content often appears fluent and confident, making it difficult for users to distinguish fabricated claims from accurately summarized information. 

\section{Data and Task}
\paragraph{Dataset.}
We use the QBias corpus \cite{haakQbiasDatasetMedia2023}, which contains 21,747 U.S. political news articles collected from AllSides balanced news headline roundups (June 2012--November 2022) and labeled by expert annotators as \emph{left}, \emph{center}, or \emph{right}. After cleaning (duplicate removal, missing text), our final set contains $N=\text{21727}$ articles with counts: left=\text{10261}, right=\text{7214}, center=\text{4252}. We retain topical tags provided by QBias (e.g., Elections, Immigration, Gun Control), enabling topic-level risk analysis. This imbalance reflects the underlying distribution of partisan sources and is accounted for through normalization in our analysis. The temporal scope (2012--2022) captures a particularly relevant period for studying political discourse, encompassing three presidential administrations, multiple election cycles, and the emergence of ``fake news'' as a prominent concern.

\paragraph{Task: news-grounded question answering.}
We evaluate \emph{grounded QA} rather than generic summarization to reduce degrees of freedom: each model must answer a question using only the provided article as evidence. For each article:
\begin{enumerate}[leftmargin=1em]
    \item \textbf{Question generation:} generate a single neutral question capturing the core claim of the article.
    \item \textbf{Grounded answering:} provide a short answer in sentence form using only the article content.
\end{enumerate}
This yields a controlled setting where sentence-level hallucination detection is well-defined.

\section{Methodology}

\begin{figure}[ht]
  \centering
  \includegraphics[width=\linewidth]{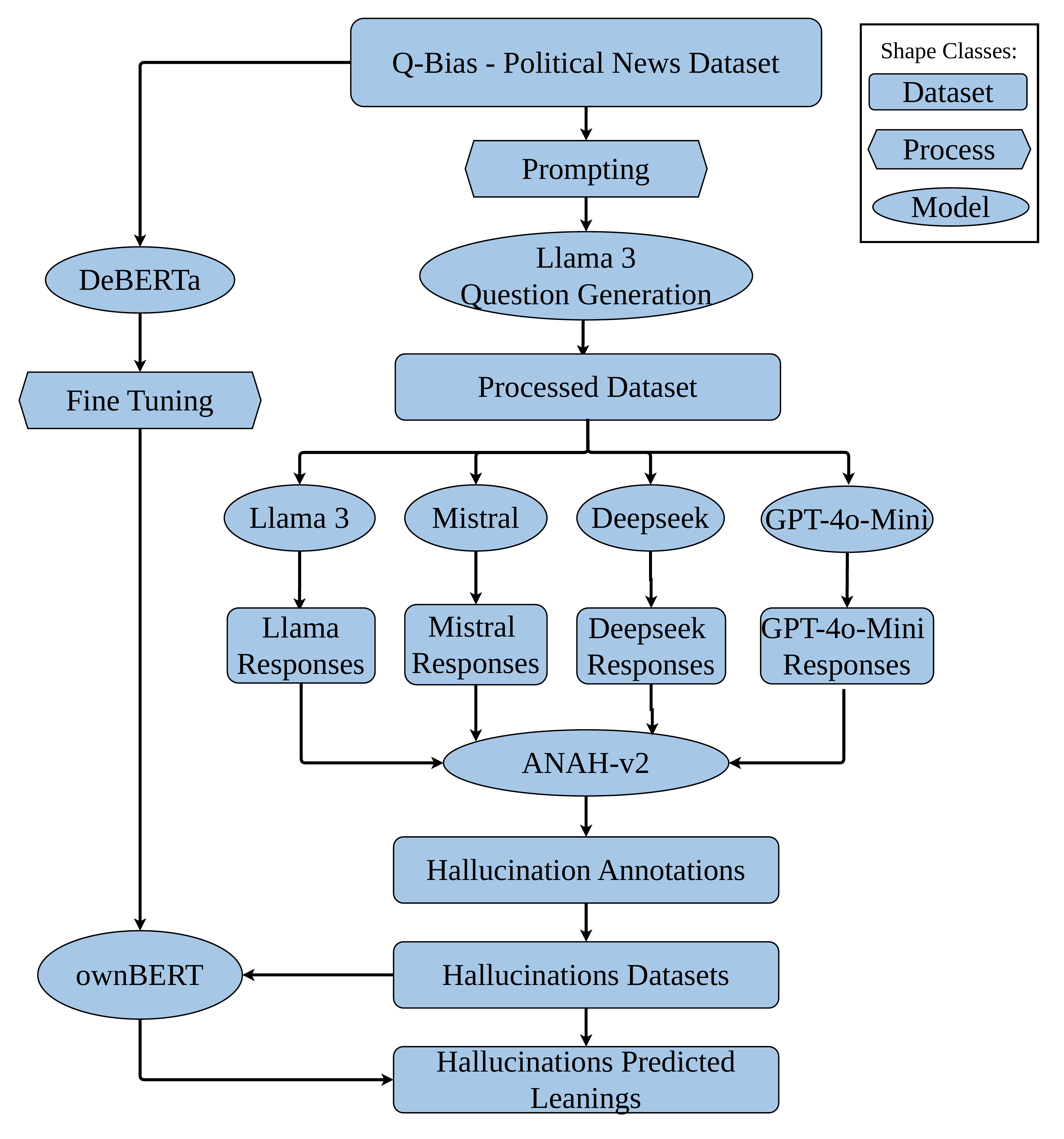}
  \caption{Pipeline overview: (1) question generation, (2) news-grounded answer generation, (3) sentence-level hallucination detection, (4) ideology classification of hallucinated sentences (binary left/right), and (5) logit-based uncertainty probing.}
  \label{fig:pipeline}
\end{figure}

Figure~\ref{fig:pipeline} illustrates the end-to-end pipeline of our methodology, from question generation to hallucination analysis and ideological assessment.

\subsection{Models and Generation}
We evaluate three instruction-tuned open-source LLMs and one proprietary LLM: \textit{Llama 3 8B Instruct}, \textit{Mistral 7B v0.3 Instruct}, \textit{Deepseek 7B Chat} and \textit{GPT-4o-Mini}. For each article, we generate a single neutral question using Llama 3 and then have each of the four models answer the same question using the full article as context (Appendix: prompts and decoding parameters). Using a single shared question per article ensures comparability across models and mitigates question-content confounds.
We segment each model answer into sentences; all hallucination and ideology analyses are conducted at the sentence level.

\subsection{Hallucination Detection (Sentence-Level)}
We apply ANAH-v2 \cite{gu2024anah} to compare each answer sentence against the corresponding source article and assigning one of three labels:
\textit{ok} (supported), \textit{unverifiable} (not supported), or \textit{contradictory} (conflicts with the article).
In all subsequent analyses, we collapse \textit{unverifiable} and \textit{contradictory} into a single hallucination indicator.
We define a binary hallucination indicator $\mathbb{H} \in \{0,1\}$ at the sentence level as
\[
\mathbb{H} \;=\; \mathbb{I}\!\left(\ell \in \{\textit{unverifiable},\, \textit{contradictory}\}\right),
\]
where $\mathbb{I}(\cdot)$ is the indicator function and $\ell$ denotes the ANAH-v2 sentence label.

\subsection{Ideology Classification of Hallucination}

To label the ideology of hallucinated content, we fine-tune DeBERTa-v3 \cite{he2021debertav3} on QBias article text for \textbf{binary} left/right classification. We adopt the binary setup because ternary (left/center/right) classification is substantially less reliable on QBias (F1 = 0.62 vs.\ 0.74), which would introduce additional noise and reduce the robustness of drift estimation. The exclusion of the ``center'' articles is therefore a methodological choice driven by classifier reliability, rather than an assumption about the underlying ideological structure.
We downsample the larger class so that the left and right training sets are balanced at $N_{\min}=\min(N_{\text{left}},N_{\text{right}})$ articles per class. The resulting classifier achieves test macro-F1 = 0.74.

We apply this classifier to hallucinated sentences and interpret its outputs as \emph{hallucination-content ideology}. Because the classifier is binary, all \textbf{directional drift analyses (RQ2)} are restricted to:
source articles labeled left or right (center excluded), and
hallucinated sentences classified as left or right (binary output).

We nevertheless include \textbf{center} sources in \textbf{RQ1} hallucination-rate and topic-risk profiling.

\subsection{Directional Drift Metrics and Hypothesis Tests}
Let $s \in \{L,R\}$ denote the ideology of source-article ideology (restricted to left/right for drift analyses) and $\hat{y}\in\{L,R\}$ denote the predicted ideology of a hallucinated sentence.

We report the following metrics:
\begin{itemize}
    \item \textit{Alignment rate.} $\Pr(\hat{y}=s \mid \text{hallucination}, s\in\{L,R\})$.
    \item \textit{Leftward drift from right sources.} $\Pr(\hat{y}=L \mid \text{hallucination}, s=R)$.
    \item \textit{Overall left-share among hallucinations.} $\Pr(\hat{y}=L \mid \text{hallucination}, s\in\{L,R\})$.
\end{itemize}

To test directionality, we conduct one-sided binomial tests of $\Pr(\hat{y}=L)$ against a symmetric null $p_0=0.5$ and report exact 95\% confidence intervals and Cohen’s $h$ effect sizes.

\subsection{Logit-Level Uncertainty Probing (RQ3)}
To investigate underlying generation mechanisms, we extract final-layer token logits during decoding and compute per-sentence uncertainty measures. 
For each token with predictive distribution $p(\cdot)$, entropy is defined as:
\[
H(p) = - \sum_{v} p(v)\log p(v).
\]
We aggregate uncertainty at the sentence level by computing the mean token entropy and include token count as a control for sentence length. We restrict the logit-level analysis to the three open-weight models: for these we compute entropy over the full output distribution, whereas GPT-4o-Mini exposes only its top-5 token log-probabilities through the API, yielding entropy that is not comparable in scale; we report a separate probe of its logits in Section~\ref{subsec:rq3}.

\noindent\textbf{T1 (uncertainty $\rightarrow$ hallucination).}
We fit logistic regression models predicting sentence-level hallucination from mean token entropy, while controlling for sentence length (token count). We report odds ratios, 95\% confidence intervals(CIs), and ROC--AUC.

\noindent\textbf{T2 (uncertainty $\rightarrow$ drift, within hallucinations).}
Restricting to hallucinated sentences, we fit logistic regressions predicting whether hallucinated content is classified as left-leaning (vs right-leaning) from mean entropy and token count. We estimate models separately by LLM to characterize heterogeneity in the uncertainty--drift relationship.
Table~\ref{tab:setup} summarizes our experimental setup in the paper. 
\begin{table}[ht]
\centering
\small
\begin{tabular}{ll}
\toprule
\textbf{Component} & \textbf{Details} \\
\midrule
Dataset & QBias (cleaned): 21,727 articles \\
Time period & June 2012 -- November 2022 \\
LLMs evaluated & Llama 3 8B, Mistral 7B, Deepseek 7B, \\
& GPT-4o-Mini \\
Hallucination detector & ANAH-v2 ($\sim$90\% accuracy) \\
Stance classifier & ownBERT (DeBERTa-v3, F1 = 0.74) \\
\bottomrule
\end{tabular}
\caption{Experimental setup summary.}
\label{tab:setup}
\end{table}

\section{Results}

\subsection{RQ1: Do Hallucination Rates Vary by Source Ideology and Topic?}

\textbf{Overall hallucination rate.} To quantify model behaviour under grounding failure, we first examine the overall volume and type of hallucinations each system produces. Understanding baseline hallucination rates is essential because it determines how much ungrounded content contributes to downstream ideological drift. 

Substantial differences emerge across models despite similar parameter counts. DeepSeek produced the highest number of hallucinations (7,090 sentences), nearly doubling the counts of Llama (3,702) and Mistral (4,475), while GPT-4o-Mini generated the fewest (1,923). This variation suggests that hallucination propensity depends not only on parameter count but also heavily on model architecture and training procedures.

The substantial variation in hallucination rates across models with similar parameter counts warrants further examination. In percentage terms, hallucination rates also differ significantly across models ($\chi^2 = 2082.20$, $p < 0.001$). Deepseek exhibits the highest hallucination rate (21.3\%), substantially exceeding Llama (14.1\%), Mistral (14.2\%) and GPT-4o-Mini (7.8\%). Pairwise comparisons reveal that Deepseek differs significantly from both Llama and Mistral ($p < 0.001$), while Llama and Mistral show identical rates ($p = 0.72$). GPT-4o-Mini also differs significantly from all three open-weight models (all $p < 0.001$), exhibiting the lowest hallucination rate. Llama 3's relatively low hallucination rate within the open-weight camp may partially reflect its use for question generation—questions it formulated may align better with its internal knowledge organization, providing a slight advantage. However, even accounting for this potential confound, Deepseek's hallucination rate represents a substantial increase over Llama, suggesting meaningful architectural or training differences beyond our experimental design. Moreover, GPT-4o-Mini's superior performance despite not benefiting from the same question-generation advantage suggests that its lower hallucination rate reflects genuine differences in model capabilities rather than an artifact of the evaluation setup. This pattern suggests that Deepseek's elevated hallucination propensity reflects a model-specific property, potentially related to training data composition or architecture, rather than task difficulty, since Llama and Mistral perform equivalently on the same inputs. Notably, all three open-weight models were instruction-tuned, yet their hallucination profiles differ markedly. This suggests that instruction tuning, while improving task compliance, does not uniformly address factual grounding. 

\noindent\textbf{By source ideology.} A natural next question is whether models' hallucination frequency varies by one ideological side. 
After normalizing for dataset imbalance, the three open-weight models exhibit hallucinations at substantially higher rates for left-leaning source articles: Deepseek (22.5\% left vs. 19.7\% center vs. 20.7\% right), Llama (15.0\% vs. 12.8\% vs. 13.4\%), and Mistral (15.1\% vs. 12.4\% vs. 14.0\%).
This statistically significant pattern ($p < 0.001$) is remarkably consistent across the three models, suggesting that left-leaning content may be systematically more challenging for them to ground accurately. In contrast, GPT-4o-Mini shows virtually no variation across source ideology (7.9\% vs. 7.7\% vs. 7.8\%; $p = 0.927$). This suggests that the ideology-linked variation in hallucination rates is concentrated among the open-weight models rather than being a universal property of the task. Potential explanations include differences in linguistic complexity, topic distribution, gaps in training data representation, or model-specific factual grounding behavior.

\begin{figure}[ht]
  \centering
  \includegraphics[trim=0 0 0 0, clip, width=0.45\textwidth]{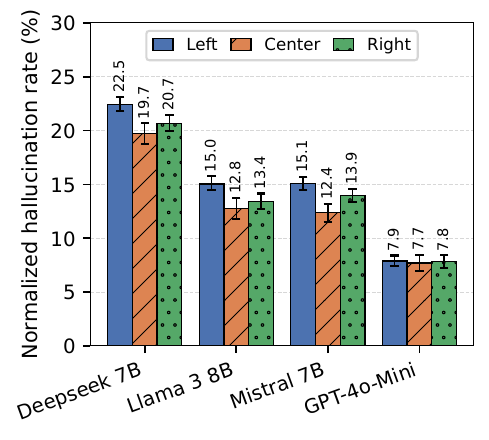}
  \caption{Normalized hallucination proportions by source leaning.}
  \label{fig:left_right_rates}
\end{figure}

\begin{figure}[t]
  \centering
  \includegraphics[trim=0 0 0 0, clip, width=0.45\textwidth]{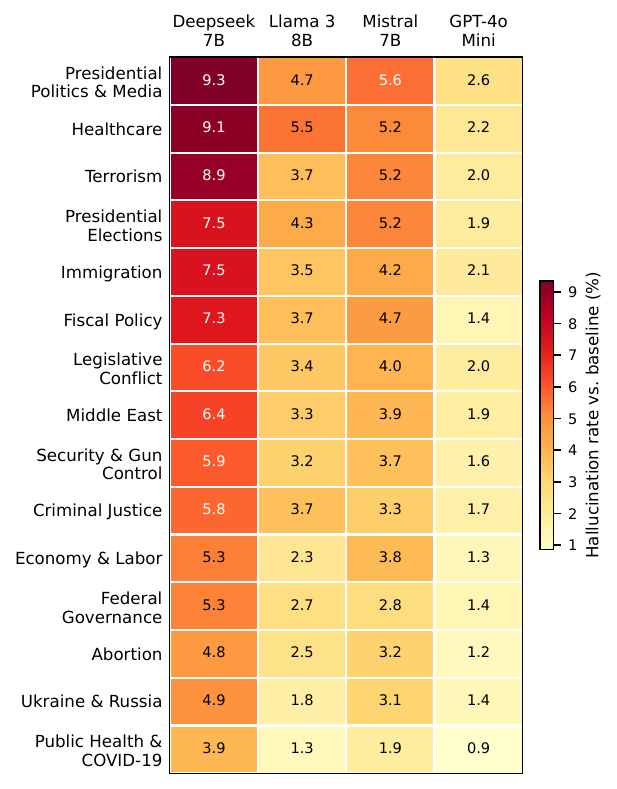}
  \caption{Hallucinated topics by model (normalized by baseline topic frequency).}
  \label{fig:topics}
\end{figure}

\noindent\textbf{By topic.} Beyond overall frequency, hallucination propensity may vary by topical domain. Political news spans diverse issue areas that differ in polarity, factual stability, and narrative complexity. To illuminate where models struggle most, we analyze hallucination rates by topic in Figure~\ref{fig:topics}.

Figure~\ref{fig:topics} shows hallucination rates by topic, normalized to baseline topic distribution in the dataset. Hallucinations cluster around contested political topics. Across all models, the top hallucination-prone topics include presidential politics, presidential elections, healthcare, immigration and gun control. These topics share common characteristics: they feature rapidly evolving information, contested factual claims, and strong partisan framing in media coverage. This suggests that training data composition plays a role---topics with abundant but conflicting information may leave models with uncertain associations, increasing hallucination likelihood.

\subsection{RQ2: Is Hallucinated Content Ideologically Directional?}

\textbf{Alignment vs. misalignment.}
After understanding where hallucinations occur, we next assess whether hallucinated content stays consistent with the ideological direction of the source article. This alignment analysis reveals how often hallucinations reinforce versus diverge from the original narrative. Because our stance classifier is binary for reliability, RQ2 drift analyses are restricted to source articles labeled left/right and hallucinated sentences classified as left/right.

\begin{figure}[ht]
  \centering
  \includegraphics[width=0.45\textwidth]{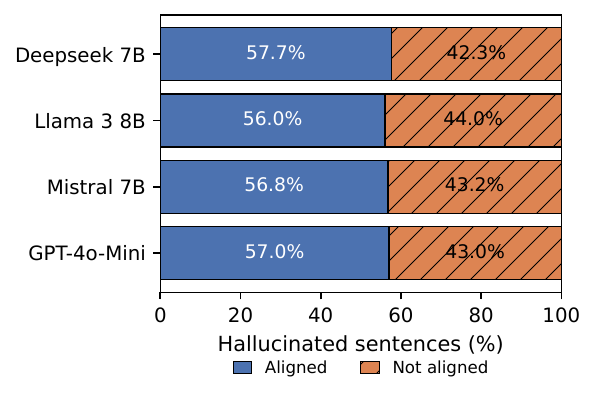}
  \caption{Ideological alignment of hallucinations: proportion matching source orientation (Aligned) vs. diverging (Not Aligned).}
  \label{fig:alignment}
\end{figure}

To quantify ideological alignment, we compare ownBERT's predicted leaning for each hallucinated sentence with the expert-annotated leaning of its source article. A sentence is considered \emph{aligned} if the labels match (left--left or right--right) and \emph{misaligned} otherwise. Figure~\ref{fig:alignment} reports the proportion of hallucinated sentences whose predicted political leaning matches the orientation of the source article.
Across all models, only 56--58\% of hallucinations preserve the ideological direction of their source---barely above chance. Approximately 43\% of fabricated content introduces perspectives \emph{absent} from the original article. Differences across models are minimal (56.0\% for LLaMA, 56.8\% for Mistral, 57.7\% for DeepSeek and 57.0\% for GPT), indicating that while hallucination \emph{frequency} varies substantially across models, ideological alignment remains largely invariant.

\noindent\textbf{Directional drift.} The alignment analysis reveals that roughly 43\% of hallucinations diverge from their source orientation, but this alone does not indicate whether drift is random or directional. The critical question is: in which direction do misaligned hallucinations drift? If errors were symmetric, we would expect roughly equal proportions of left/right misclassifications. Systematic asymmetry, by contrast, would indicate an underlying ideological bias in ungrounded generation. To measure the directionality of misaligned hallucinations, we compare the predicted ideological leaning of each hallucination against its source article. 

\begin{figure}[ht!]
    \centering
    {\small\textbf{(a) \textsc{Llama}}}\\
    \includegraphics[width=0.65\linewidth]{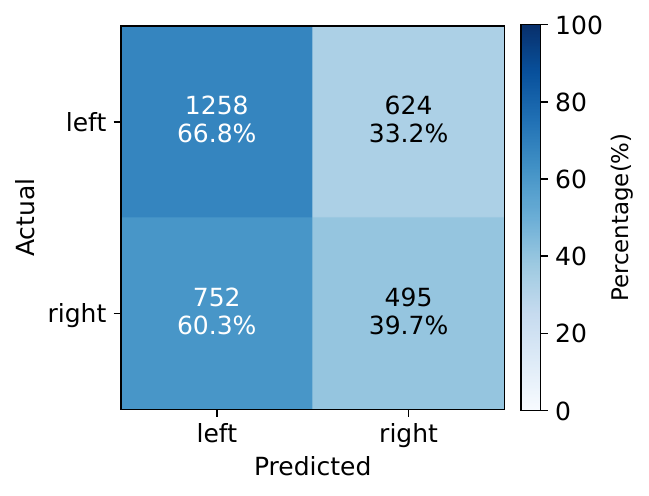}
    
    {\small\textbf{(b) \textsc{Mistral}}}\\
    \includegraphics[width=0.65\linewidth]{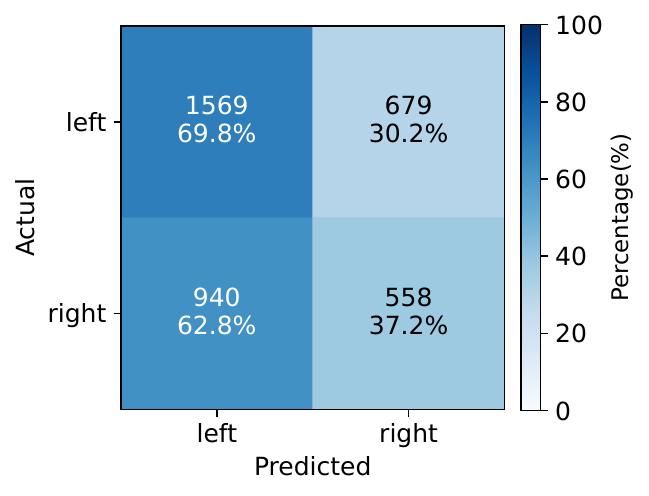}
    
    {\small\textbf{(c) \textsc{Deepseek}}}\\
    \includegraphics[width=0.65\linewidth]{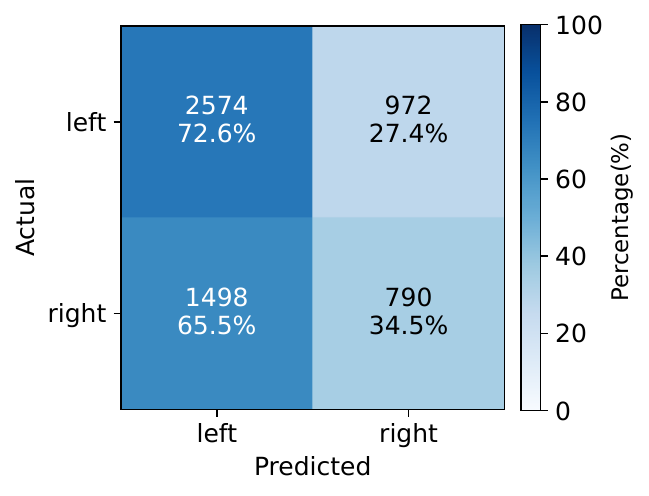}

      {\small\textbf{(d) \textsc{GPT-4o-Mini}}}\\
    \includegraphics[width=0.65\linewidth]{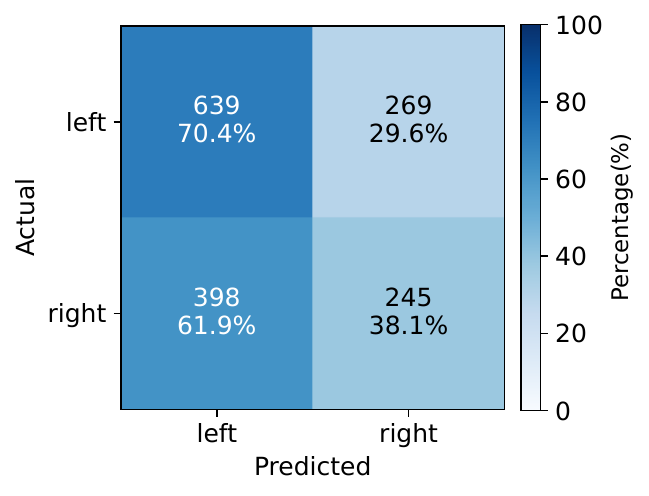}
  \caption{Confusion matrices: source article orientation (rows) vs. predicted hallucination orientation (columns).}
  \label{fig:confusion}
\end{figure}

Figure~\ref{fig:confusion} shows the confusion matrices comparing source article orientation (rows) with predicted hallucination orientation (columns). All four models exhibit systematic leftward drift: regardless of source ideology, hallucinated content is predominantly classified as left-leaning. Examining the overall proportion of left-leaning predictions among all predictions: Deepseek shows the strongest skew (69.8\% of all hallucinations classified as left), followed by Mistral (67.0\%), GPT-4o-Mini (66.9\%) and Llama (64.2\%).

Critically, this leftward skew persists even when doing news-grounded QA of right-leaning articles. For right-leaning sources, the proportion of hallucinations classified as left-leaning is: Deepseek 65.5\%, Mistral 62.8\%, GPT-4o-Mini (61.9\%) and Llama 60.3\%. This asymmetry confirms that hallucinations do not introduce random noise, as they systematically inject left-leaning content regardless of source orientation. The consistency of this pattern across four independently developed models suggests a common underlying cause, likely related to the composition of English-language training corpora.

Given the frequency asymmetry above, one might hypothesize that hallucination political bias simply reflects the higher volume of hallucinations from left-leaning sources. However, this explanation fails: even when models hallucinate on right-leaning source articles, the hallucinated content still drifts leftward. This demonstrates that leftward content bias is independent of source ideology, and the models inject left-leaning content regardless of what they are reading. Table ~\ref{tab:drift}  presents binomial test results.

\begin{table}[tb]
\centering
\small
\setlength{\tabcolsep}{4pt} 
\begin{tabular}{lcccc}
\toprule
\textbf{Condition} & \textbf{\% Left} & \textbf{95\% CI} & \textbf{$p$-value} & \textbf{$h$} \\
\midrule
Overall
& 67.5
& [66.7, 68.3]
& $<0.001$
& 0.36 \\
\midrule
Deepseek 7B Chat
& 69.8
& [68.6, 71.0]
& $<0.001$
& 0.41 \\
Mistral 7B v0.3 Instruct
& 67.0
& [65.4, 68.5]
& $<0.001$
& 0.35 \\
Llama 3 8B Instruct
& 64.2
& [62.5, 65.9]
& $<0.001$
& 0.29 \\
GPT-4o-Mini
& 66.9
& [64.5, 69.2]
& $<0.001$
& 0.34 \\
\midrule
Right sources only
& 63.2
& [61.9, 64.5]
& $<0.001$
& 0.27 \\
\bottomrule
\end{tabular}
\caption{
Leftward drift in hallucinated outputs. Entries report the proportion of left-leaning predictions among hallucinated responses classified as either left or right. $p$-values are from one-sided binomial tests against a null of equal left/right probability ($p_0 = 0.5$). Models show drift with medium effect sizes (Cohen's $h > 0.25$). 
}
\label{tab:drift}
\end{table}

If question generation confounded results by biasing content toward Llama's priors, Llama should exhibit the strongest leftward drift. Instead, Llama shows the \textit{weakest} drift (64.2\% vs.\ Deepseek's 69.8\%; $\chi^2 = 30.51$, $p < 0.001$). Furthermore, Llama and Mistral show statistically identical hallucination rates (14.1\% vs.\ 14.2\%, $p = 0.72$), suggesting no unique Llama advantage. These patterns argue against the QG confound driving our findings.

\subsection{RQ3: Do Uncertainty Patterns Explain Hallucination and Drift?}\label{subsec:rq3}

\textbf{T1: Uncertainty predicts hallucination.} Logit-level analysis indicates that hallucinations arise in high-uncertainty generation contexts. Mean token entropy is substantially higher for hallucinated sentences ($M = 0.639$) than for correct sentences ($M = 0.323$), yielding a difference that is both statistically significant (Welch's $t = 100.3$, $p < 0.001$) and practically large (Cohen's $d = 1.03$). 

To quantify the predictive relationship, we fit a logistic regression predicting hallucination from mean entropy while controlling for response length (number of tokens). Each unit increase in mean entropy multiplies hallucination odds by 40.4 (95\% CI: [37.6, 43.3], $p < 0.001$). The model achieves ROC-AUC = 0.78, indicating good discriminative ability. This supports an ``uncertainty $\rightarrow$ guessing'' mechanism consistent with \cite{kalai2026evaluating}.

\noindent\textbf{T2: Uncertainty-ideology relationship.} We also examined whether higher uncertainty predicts leftward drift within hallucinations, using logistic regression to control for response length. Models differ markedly:

Deepseek shows a strong uncertainty-ideology relationship: each unit increase in mean entropy multiplies the odds of left-leaning hallucination content by 2.32 (95\% CI: [2.02, 2.67], $p < 0.001$). Leftward bias emerges specifically in high-uncertainty contexts. Llama shows a moderate relationship in the same direction (OR = 1.47, 95\% CI: [1.14, 1.89], $p = 0.003$). Mistral shows no significant uncertainty-ideology relationship (OR = 0.71, 95\% CI: [0.50, 1.02], $p = 0.065$), indicating that Mistral's leftward bias operates independently of model confidence, i.e. the bias is present regardless of uncertainty level.

This heterogeneity suggests different bias mechanisms: Deepseek and Llama's biases activate under uncertainty, while Mistral's bias appears more uniformly embedded. This indicates that while all models share leftward-biased priors, the mechanism differs across models.

\noindent\textbf{Mechanistic interpretation.} Our results adjudicate between two hypotheses about how leftward bias manifests. Under H1 (uncertainty-gated priors), leftward drift becomes more likely as uncertainty increases. Under H2 (uncertainty-invariant priors), drift is present regardless of uncertainty level. Our results support H1 for Deepseek and Llama: within hallucinations, higher entropy significantly increases the odds that hallucinated content is classified as left-leaning. In contrast, Mistral shows no positive uncertainty--drift relationship, consistent with H2 or with mechanisms not captured by entropy alone, suggesting a more uniformly embedded prior that neither hypothesis fully explains. We therefore interpret uncertainty as a robust trigger for hallucination, and (for some models) a trigger for directional drift; identifying the source of leftward priors (e.g., training data composition or alignment) remains outside the scope of these analyses.

This heterogeneity implies that bias mitigation strategies may need to be model-specific: uncertainty detection may help for Deepseek and Llama, while Mistral may require interventions at the training or architectural level. These findings also explain existing observations that AI produces more errors when advocating right-leaning positions~\cite{lin2025persuading} generating right-leaning content requires overcoming the model's latent leftward prior, increasing error susceptibility. While our results illuminate the \textit{trigger mechanism}, they do not identify the \textit{source} of leftward priors, which likely reflects training data composition.

\noindent\textbf{Closed-model probe.} We also applied the logit analysis to GPT-4o-Mini, but its API exposes only the top-5 token log-probabilities. The resulting entropy is sharply compressed with max sentence-level entropy 0.70 nats versus 3.3--3.4 for the open-weight models, and the separation between hallucinated and correct sentences nearly vanishes (0.04 nats vs.\ 0.33--0.38), leaving its uncertainty--drift estimate uninformative (OR = 1.58, 95\% CI: [0.62, 4.03], $p = 0.343$). We therefore exclude GPT-4o-Mini from the RQ3 estimates above. This limitation is itself relevant to auditing: a deployed proprietary model's uncertainty mechanism cannot be probed this way without fuller access to its output distribution.

\section{Discussion}

\noindent\textbf{Directional hallucinations as structured errors.}
Our results separate two phenomena that are often conflated:
\begin{itemize}[leftmargin=1em]
    \item \textbf{Reliability (how often models hallucinate):} models hallucinate more frequently on left-leaning sources, but the difference is not large across left/center/right source material in our grounded QA setup.
    \item \textbf{Ideological structure of errors (what they hallucinate):} systematically left-skewed hallucination content, even when the source is right-leaning. This finding aligns with documented left-leaning tendencies in LLM outputs~\cite{rozadoMeasuringPoliticalPreferences2025,linTrustworthyLLMsReview2024,chen2026uncovering}.
\end{itemize}
This dissociation matters for social-good deployments: even if overall hallucination rates are reduced, the remaining errors may still be directionally biased.

\medskip
\noindent\textbf{Mechanisms: uncertainty, guessing incentives, and priors.}
Logit-level evidence supports that hallucinations concentrate in high-uncertainty contexts, aligning with the argument that current training/evaluation regimes reward ``guessing'' \cite{kalai2026evaluating}. Yet leftward drift persists even when uncertainty patterns differ by model, consistent with the view that hallucinations expose asymmetric priors embedded during pretraining and alignment (e.g., corpus composition, RLHF preferences, or safety tuning) \cite{navigli2023biases,gallegos2024bias,santurkarWhoseOpinionsLanguage2023}.

\medskip
\noindent\textbf{Election-specific implications.}
In election-adjacent contexts, both \emph{factual error} and \emph{perceived partisan skew} can reduce trust, even when systems are intended to be neutral. Our topic-level profiling provides a practical pathway: models can be audited and constrained more aggressively on election-relevant topics (e.g., impose abstention at high uncertainty, require citations, enforce retrieval grounding) \cite{geifman2017selective,guo2017calibration,lewis2020retrieval}.

\section{Limitations}

\textbf{Classifier limitations.} Our classifier (F1 = 0.74) may introduce noise, but random errors bias toward null, making our significant findings conservative. We cannot fully rule out differential accuracy across ideologies; future work should examine classifier calibration by ideology.

\noindent\textbf{Data contamination.} QBias articles (2012--2022) may appear in training corpora. However, memorization would \textit{reduce} hallucination, not bias direction. Consistent leftward drift across models with different training data (including Deepseek) suggests contamination is not the primary driver.

\noindent\textbf{Detection validity.} ANAH-v2 was not validated specifically on political news. While ANAH-v2 was validated on data containing substantial political content (37\% Politics/Military), our U.S. partisan news sources may differ in style from the validation corpus.

\noindent\textbf{Question generation confound.} Using one model for question generation may advantage that model in answering. If Llama-generated questions biased content leftward, Llama should show the strongest effect. Instead, Llama shows the \textit{weakest} drift (64.2\%), arguing against this confound.

\noindent\textbf{Logit access for closed models.} Our logit-level uncertainty analysis (RQ3) requires access to the full output distribution, which closed APIs do not expose; as quantified by the closed-model probe in Section~\ref{subsec:rq3}, this precludes a comparable analysis for GPT-4o-Mini, so RQ3 is restricted to the open-weight models. GPT-4o-Mini's RQ1 and RQ2 results, which do not rely on logits, are unaffected.

\section{Implications for AI-Mediated Information Systems}

These findings have broad implications for the design, deployment, and governance of AI systems that mediate political information:

\noindent\textbf{For users and society}: As LLMs become integrated into news consumption workflows, users may unconsciously internalize politically biased misinformation. Even when source material is balanced or right-leaning, hallucinations can inject left-leaning perspectives, potentially reinforcing echo chambers and contributing to societal polarization---particularly concerning given assumptions of AI neutrality. Users who rely on AI summaries without consulting original sources may develop systematically skewed understandings of political issues.

\noindent\textbf{For platform design}: AI-powered QA systems should implement hallucination-aware content warnings and provide transparency about model tendencies. Factual grounding through retrieval-augmented generation~\cite{niuRAGTruthHallucinationCorpus2024} may help, but our results suggest that when grounding fails, systematic biases emerge. Platforms might consider displaying confidence indicators or providing links to source material for verification.

\noindent\textbf{For AI governance}: 
Policymakers should consider establishing accuracy and neutrality standards for AI systems used in politically sensitive contexts. Our methodology provides a template for auditing political bias in AI-generated content that could be applied to new models as they are released. Recent work has also revealed linguistic bias in governmental documents \cite{DBLP:conf/www/SwartHC25}, suggesting that bias can emerge in both source texts and AI-generated content.

\noindent\textbf{For model development}: Bias mitigation efforts should be topic-aware, as our analysis reveals that contested issues amplify partisan biases. Simply reducing overall hallucination rates may not address the systematic ideological skew in remaining hallucinations. Developers might consider balanced training data curation or targeted fine-tuning on politically sensitive topics.

\section*{Ethical Statement}
This research relies exclusively on publicly available, annotated news records. The dataset contains no personally identifiable information and does not involve private individuals. No human participants were recruited or directly engaged in the study; consequently, informed consent was not required.

\section*{Acknowledgements}

This work was supported by the Dutch Research Council (NWO) through the project \textit{PoliBiasEU: A Scalable and Multilingual Benchmark for Political Bias Detection in Large Language Models} (project no. ZZKNS44563) and by the Dutch Ministry of Education, Culture and Science (OCW) in the scope of the project \textit{Navigating the Storm: European Political Contestation in Geopolitical Transformation (NEST)}, and partially supported by the Network Institute, Vrije Universiteit Amsterdam, through the Academy Assistants Program.



\bibliographystyle{named}
\bibliography{ijcai26}

\clearpage

\section*{Appendix}

\section{Data example}

\begin{table}[!htbp]
\centering
\caption{Example Row From the QBias Dataset}
\label{tab:qbias}

\small
\begin{tabularx}{\columnwidth}{|c|X|}
\hline
\textbf{id} & 21629 \\ \hline
\textbf{title} & How the White House is Handling Brittney Griner's Russia Detention \\ \hline
\textbf{tags} & [World, White House, Russia, Brittney Griner] \\ \hline
\textbf{heading} & Biden under fire for handling of Brittney Griner’s detention in Russia \\ \hline
\textbf{source} & Washington Examiner \\ \hline
\textbf{text} &
 The Biden administration is facing intense scrutiny over its handling of WNBA star Brittney Griner's detention in Russia, with some questioning the White House's commitment to helping based on Griner's race and gender. Detained in Russia since Feb. 17 on marijuana-related charges, Griner has penned a letter to President Joe Biden while her friends and family plead with the administration to help. The State Department deemed her wrongfully detained in May, and Biden met with her wife, Cherelle, on Wednesday. The president offered his support to Cherelle and Brittney's family, and he committed to ensuring they... \\ \hline
\textbf{bias\_rating} & right \\ \hline
\end{tabularx}
\end{table}

\section{Models and Generation}

We evaluate three instruction-tuned open-weight LLMs---Llama 3 8B Instruct, Mistral 7B v0.3 Instruct, and Deepseek 7B Chat---together with one proprietary model, GPT-4o-Mini, included as a closed-source reference point. The open-weight models were selected based on three criteria: (1) open accessibility enabling reproducible research, (2) comparable parameter counts (7--8B) allowing fair comparison, and (3) demonstrated instruction-following capabilities. They represent current capabilities among freely available systems~\cite{madaanQuantifyingVarianceEvaluation2024} while remaining computationally feasible. All four models are used in the behavioral analyses (RQ1 and RQ2); the logit-level analysis (RQ3) is restricted to the three open-weight models, since GPT-4o-Mini exposes only its top-5 token log-probabilities through the API.

Our generation procedure follows a two-step process. First, we generate a focused question for each article using Llama 3, selected for its top ranking on the Open LLM Leaderboard~\cite{OpenLLMLeaderboard2025}.
The question-generation prompt instructs the model to formulate a single question that captures the article’s core argument. The exact prompt used is provided below.
\vspace{1mm}


\begin{quote}
\textit{
You are an expert at distilling complex news articles into concise, insightful questions.
Your goal is to generate a *single*, focused question that captures the absolute core point or main argument of the provided news article.
Here are the key rules to follow:
1.  **Be Singular:** Generate only one question.
2.  **Be Focused:** The question must directly address the central theme or most important takeaway of the article, not a minor detail.
 3.  **Be Comprehensive (of the core):** The question should encapsulate the primary subject and key action/event/implication discussed.
4.  **Avoid Personal Opinions/Bias:** Do not inject your own opinions or biases into the question.
5.  **Avoid Leading Questions:** The question should be neutral and not presuppose a specific answer.
 6.  **Be Concise:** Aim for clarity and brevity. Avoid overly long or convoluted phrasing.
 7.  **No Explanations or Context:** Do not provide any introductory phrases, explanations, or additional context outside of the question itself. Just the question.
8.  **Directly Actionable:** The question should prompt a discussion or reflection on the article's core message.
Bad Example of a question (too broad, not specific enough to *this* article):
 	- "What is happening in the world?"
Good Example of a question (specific to an article about a new climate policy):
- "How will the proposed carbon tax impact the nation's energy sector and consumer costs?"
}
\end{quote}

Second, each of the four LLMs answers these questions using the full article as context. The news-grounded QA prompt instructs the models to provide factual, sentence-level answers grounded exclusively in the source text. This setup creates a controlled environment in which hallucinations—content not supported by the article—can be clearly detected. The exact prompt used is provided below.

\begin{quote}
\noindent\textit{
"You are an expert at extracting specific information from news articles to answer questions directly and concisely. "
"For each question, you MUST provide a factual answer based on the accompanying article. "
"Provide your answer in sentence format. Do NOT use bullet points, numbered lists, or any other list format. "
"DO NOT elaborate, ask clarifying questions, explain your reasoning, or list steps. "
"Provide ONLY the answer."
}
\end{quote}

\section{Ideological Classifier Fine-Tuning and Validation}

To ensure that the observed ideological skew in hallucinated outputs is not an artifact of the ideological classifier itself, we fine-tune and validate an independent left--right text classifier based on a pretrained transformer model.

\subsection{Training Data and Preprocessing}

We construct a balanced binary classification dataset consisting of news articles labeled as \textit{left}- or \textit{right}-leaning. To eliminate class imbalance as a potential source of bias, we downsample the majority class such that the final dataset contains an equal number of left- and right-labeled observations (7,226 per class). The dataset is split into training (80\%), validation (10\%), and test (10\%) sets using stratified sampling, preserving class balance across all splits. The held-out test set contains 1,446 observations (723 per class).

Each input consists of the article headline concatenated with the article body text. Texts are tokenized using the DeBERTa-v3 tokenizer and truncated to a maximum sequence length of 256 tokens.

\subsection{Model Architecture and Training Procedure} \label{sec:finetune} 

We fine-tune \texttt{microsoft/deberta-v3-base} for binary sequence classification. The classification head is randomly initialized and trained end-to-end. Model training is performed for five epochs using the AdamW optimizer with a learning rate of $2 \times 10^{-5}$ and weight decay of 0.01. Training uses a batch size of 16 and evaluation uses a batch size of 32. Mixed-precision training (FP16) is enabled when supported by the hardware.

To further guard against residual imbalance, we employ inverse-frequency class weighting in the cross-entropy loss function. Model selection is based on macro-averaged F1 score on the validation set, and the best-performing checkpoint is retained for evaluation.

\subsection{Classifier Performance}

\begin{figure}[ht]
  \centering
  \includegraphics[width=0.4\textwidth]{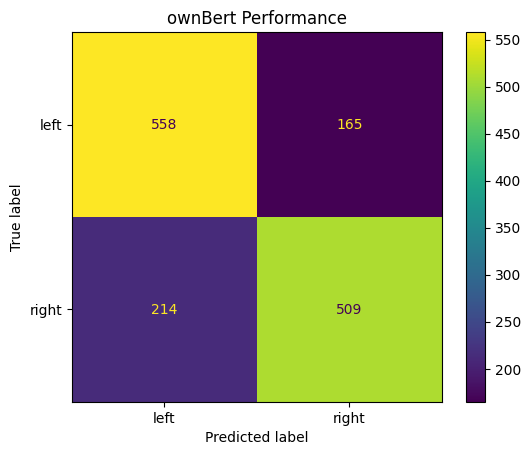}
\caption{Confusion matrix for the binary stance classifier on the held-out test set.}
\label{fig:classifier_cm}
\end{figure}

On the held-out test set, the classifier achieves an overall accuracy of 0.74 and a macro-averaged F1 score of 0.74, indicating balanced performance across ideological classes. Figure ~\ref{fig:classifier_cm} reports the confusion matrix. This implies recall of 0.77 for left-labeled articles and 0.70 for right-labeled articles.

Across the entire test set, the classifier predicts \textit{left}-leaning labels in 53.4\% of cases, reflecting a mild asymmetry that is substantially smaller than the 64--70\% leftward skew observed in hallucinated outputs produced by large language models. This gap indicates that the classifier itself cannot account for the magnitude of ideological drift identified in the main analysis.

\subsection{Implications for Bias Measurement}

These results confirm that the ideological classifier is not the primary driver of the leftward skew observed in hallucinated content. The classifier is trained on balanced data, exhibits comparable performance across ideological classes, and shows only a modest tendency toward left predictions. Consequently, the pronounced leftward drift documented in hallucinated outputs reflects properties of the generative models rather than artifacts of downstream classification.

\end{document}